\documentclass[11pt,a4paper]{article}
\usepackage{authblk}
\usepackage[hyperref]{acl2021}
\usepackage{times}
\usepackage{latexsym}
\usepackage{url}

\usepackage{array,multirow}
\usepackage{graphicx}
\usepackage{bm}
\usepackage{amsmath}
\usepackage{amsfonts}
\usepackage{url}
\usepackage{nidanfloat}
\usepackage{algorithmic, algorithm}
\usepackage{comment}
\usepackage{CJKutf8}

\renewcommand{\b}{\textbf}
\usepackage{microtype}

\usepackage{subfig}
\usepackage[format=plain,font=small]{caption}

\usepackage{microtype}
\aclfinalcopy 

\title{
    Joint Optimization of Tokenization and Downstream Model
    }

\author{
  \textbf{Tatsuya Hiraoka}$^\dagger$,
  \textbf{Sho Takase}$^\dagger$,
  \textbf{Kei Uchiumi}$^\ddagger$,
  \textbf{Atsushi Keyaki}$^\ddagger$,
  \textbf{Naoaki Okazaki}$^\dagger$
  \\\\
  $^\dagger$ Tokyo Institute of Technology\\
  $^\ddagger$ Denso IT Laboratory, Inc.\\
  \texttt{\{tatsuya.hiraoka, sho.takase\}@nlp.c.titech.ac.jp} \\
  \texttt{\{kuchiumi, akeyaki\}@d-itlab.co.jp} \\
  \texttt{okazaki@c.titech.ac.jp}
  }

\date{}

\begin{document}
\maketitle
\begin{abstract}
Since traditional tokenizers are isolated from a downstream task and model, they cannot output an appropriate tokenization depending on the task and model, although recent studies imply that the appropriate tokenization improves the performance.
In this paper, we propose a novel method to find an appropriate tokenization to a given downstream model by jointly optimizing a tokenizer and the model.
The proposed method has no restriction except for using loss values computed by the downstream model to train the tokenizer, and thus, we can apply the proposed method to any NLP task.
Moreover, the proposed method can be used to explore the appropriate tokenization for an already trained model as post-processing.
Therefore, the proposed method is applicable to various situations.
We evaluated whether our method contributes to improving performance on text classification in three languages and machine translation in eight language pairs.
Experimental results show that our proposed method improves the performance by determining appropriate tokenizations.
\end{abstract}

\section{Introduction}
\label{sec:intro}
Tokenization, which converts a raw sentence into a sequence of tokens, is a crucial process that affects the performance of NLP tasks.
Existing studies have proposed various tokenization methods including rule-based tokenization~\cite{koehn2007moses}, dictionary-based tokenization~\cite{kudo2006mecab,morita-etal-2015-morphological,tolmachev-etal-2018-juman,takaoka18sudachi}, supervised tokenization with neural networks~\cite{yang2017neural, cai2017fast, yang2018subword}, and unsupervised tokenization~\cite{goldwater2006contextual, goldwater2009bayesian, mochihashi2009bayesian, sennrich2016neural,kudo2018sentencepiece}.
Much of prior research has reported that an appropriate tokenization depends on each downstream task~\cite{xu2008bayesian, chang2008optimizing, nguyen2010nonparametric, domingo2018much, hiraoka2019stochastic, gowda2020finding}.
Moreover, \newcite{hiraoka2020optimizing} implies that we have to consider a downstream model to determine an appropriate tokenization.
In other words, we can improve the performance of a downstream model by determining an appropriate tokenization for the downstream model.
However, since traditional tokenizers are isolated from a downstream model, we need to train a given downstream model with each possible tokenization and evaluate its performance to determine the appropriate tokenization.
Performing such an exploration whenever we construct a new downstream model is impractical.

Several studies have addressed the optimization of a tokenizer based on a downstream task or/and model~\cite{xuanli2020dynamic,hiraoka2020optimizing}, but existing methods are restricted to specific tasks.
\newcite{xuanli2020dynamic} proposed DPE as a tokenization method for a sequence-to-sequence problem such as machine translation.
Their method trains a tokenizer with a given training corpus, but it is isolated from a downstream model such as a neural encoder--decoder for machine translation.
\newcite{hiraoka2020optimizing} proposed OpTok, which jointly trains a tokenizer and a downstream model.
However, its architecture is specific to classification problems based on sentence representations, and thus, it cannot be applied for various tasks such as sequence-to-sequence problems.
Therefore, there is no method to optimize a tokenizer depending on any downstream task and model.

In this paper, we propose a novel method to jointly optimize a tokenizer and downstream model without any restriction on a task\footnote{Code: \url{https://github.com/tatHi/optok4at}}.
The proposed method can determine an appropriate tokenization for a downstream model because it explores different tokenizations based on loss values of the downstream model.
Since the proposed method requires only loss values of the downstream model, we can apply it for any task and model.
Moreover, even if a given downstream model is already trained, our proposed method can be applied to improve the performance by refining tokenization.
We call this refinement of tokenization {\it post-processing}.
Thus, we can easily use the proposed method in various situations including the case where we have a sufficiently trained downstream model.

We conducted experiments on text classification and machine translation tasks in various languages.
Experimental results indicate that the proposed method outperformed existing tokenization methods in both the tasks.
We also showed that our method can enhance the performance by refining tokenization as post-processing for downstream models trained with subword regularization~\cite{kudo2018subword, provilkov2019bpe}.

\section{Proposed Method}
\label{sec:proposed_method}
The proposed method comprises a tokenizer and a downstream model.
We optimize the two modules simultaneously.
First, we present the training outline of the case where we use one sentence as an input in Section \ref{sec:optimizing_tokenization}.
Second, we introduce the training of the tokenizer (Section \ref{sec:nulm_tokenizer}) and the downstream model (Section \ref{sec:model_loss}).
Finally, we explain the training strategy for a task that requires multiple inputs such as machine translation (Section \ref{sec:multi_sentences}).

\begin{figure}[t]
\centering
\includegraphics[scale=0.6]{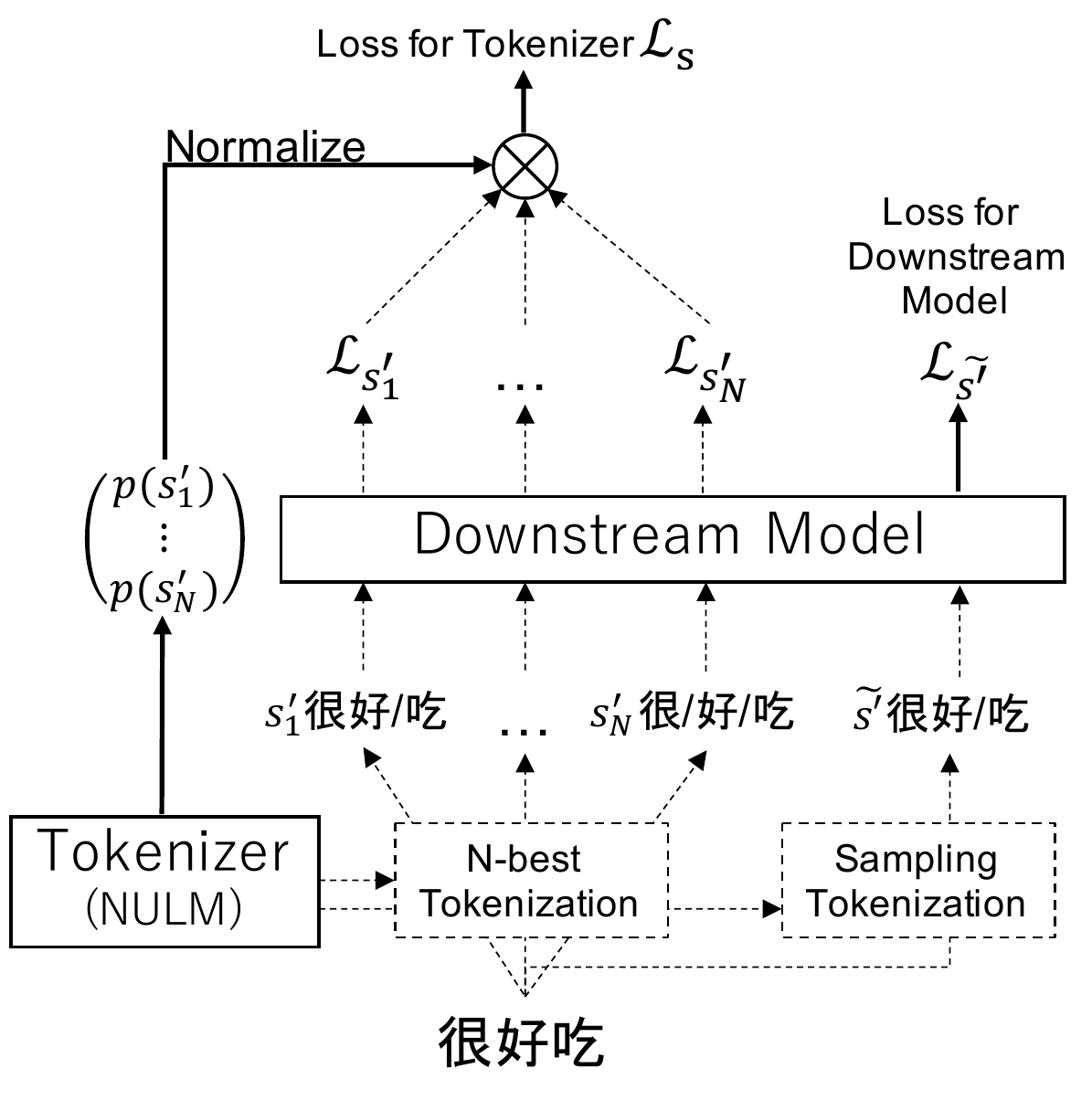}
\caption{
    Overview of the proposed method calculating a loss for a tokenizer $\mathcal{L}_s$ and that for a downstream model $\mathcal{L}_{\tilde{s'}}$ using $N$-best tokenizations (Section \ref{sec:nulm_tokenizer}) and a sampled tokenization (Section \ref{sec:model_loss}), respectively.
    The arrows with the continuous lines indicate the differentiable path for back-propagation.
}
\label{fgr:model_outline}
\end{figure}

\subsection{Optimizing Tokenization with Loss}
\label{sec:optimizing_tokenization}
The proposed method tokenizes a sentence $s$ into a sequence of words $w$ in vocabulary $V$, $s'=w_1,...,w_I$, where $I$ is the sequence length.
In this tokenization process, our purpose is to minimize the following loss value:
\begin{align}
{\mathcal L}_{s'} &= q(f(s'), z), \label{eq:nulm_loss}
\end{align}
where $f(s')$ is a downstream model that outputs a prediction of the downstream task from a tokenized sentence $s'$, and $q(f(\cdot), z)$ is a task-specific loss function between a model prediction and supervisory signal $z$.

Figure \ref{fgr:model_outline} presents an outline of the proposed method.
To determine the tokenization satisfying ${\rm argmin}_{s'}(q(f(s'), z))$, we update the tokenizer to assign a higher probability to a useful tokenization for the downstream model.
Concretely, we construct $N$ tokenizations $s'_1,...,s'_n,...s'_N$ for a training instance and then compute loss values for each tokenization.
We weight each loss based on probability $p(s'_n)$ computed by the tokenizer and use the weighted sum to train the tokenizer, as follows:
\begin{align}
a_n &= \frac{p(s'_n)}{\sum_{m=1}^{N}{p(s'_m)}}, \label{eq:weight}\\
{\mathcal L}_s &= \sum_{n=1}^{N}{a_n{\mathcal L}_{s'_n}}. \label{eq:total_loss}
\end{align}
In this study, we used $N$-best tokenizations.
In these equations, we weight losses ${\mathcal L}_{s'_1},...,{\mathcal L}_{s'_N}$ corresponding to $N$-best tokenizations with their sentence probabilities normalized such that the sum is 1.
By optimizing the tokenizer based on the weighted sum, ${\mathcal L}_s$, the tokenizer assigns high probability to the appropriate tokenization for the downstream model.

We can use any function for $f(\cdot)$ and $q(f(\cdot), \cdot)$ in Eq.(\ref{eq:nulm_loss}).
Therefore, the proposed method has no restrictions on the downstream task and model.
For instance, in the case where text classification is the downstream task, $f(\cdot)$ is a neural network predicting a label of a given tokenized sentence, and $q(f(\cdot), \cdot)$ is the cross-entropy loss between the model prediction and the true label.

\subsection{Tokenizer: NULM}
\label{sec:nulm_tokenizer}
We employ a neural unigram language model (NULM) as our tokenizer.
It calculates the unigram probability of a word $p(w)$ with a word embedding ${\bm v}_w$ as follows:
\begin{align}
d_w &= \mathrm{MLP}({\bm v}_w),  \label{eq:mlp_in_langmodel} \\
p(w) &= \frac{\mathrm{exp}(d_w)}{\sum_{\hat w \in V}\mathrm{exp}(d_{\hat w})}, \label{eq:token_prob}
\end{align}
where ${\rm MLP}(\cdot)$ is a multilayer perceptron.
We initialize vocabulary $V$ with a reasonable size of words.
For example, we can use a tokenization by SentencePiece~\cite{kudo2018sentencepiece} or BPE~\cite{sennrich2016neural} for the initialization.
We also calculate the probability of a tokenization $p(s')$ as follows:
\begin{align}
p(s')&=\prod_{w \in s'}p(w).
\end{align}

For the training with Eq.(\ref{eq:total_loss}), we obtain $N$-best tokenizations by applying Forward-DP Backward-A* algorithm~\cite{nagata1994stochastic} for possible tokens against sentence $s$.
In the inference phase, we can also obtain the $1$-best appropriate tokenization for the downstream task using Viterbi algorithm~\cite{viterbi1967error} for the trained NULM.

\subsection{Downstream Model Training}
\label{sec:model_loss}
We can train the downstream model with loss ${\mathcal L}_s$ in Eq.(\ref{eq:total_loss}), but we use subword regularization~\cite{kudo2018subword} to obtain a better model.
Thus, we compute ${\mathcal L}_{\tilde{s'}} = q(f(\tilde{s'}), z)$ for a sampled tokenization $\tilde{s'}$ and use ${\mathcal L}_{\tilde{s'}}$ to train the downstream model.

We sample tokenization $\tilde{s'}$ from $p(\tilde{s'})^\alpha/\sum_{k=1}^{K}{p(s'_k)^\alpha}$ computed by the NULM in Eq.(\ref{eq:token_prob})~\cite{kudo2018subword}.
Here, $\alpha \in \mathbb{R}^+$ is a hyperparameter that controls the diversity of the sampled tokenization.
If we set $\alpha$ as a lower value, the distribution is similar to the uniform distribution; otherwise, the distribution strongly depends on each tokenization probability $p(\tilde{s'})$.
$K$ is also a hyperparameter denoting the number of candidates for sampling, and we use Forward Filtering Backward Sampling~\cite{scott2002bayesian,mochihashi2009bayesian} if $K=\infty$.

Subwod regularization not only sophisticates the downstream model but also provides various tokenizations to the downstream model during training.
Therefore, subwod regularization helps in exploring the appropriate tokenization.

\subsection{Training in Multiple Sentences as Inputs}
\label{sec:multi_sentences}
Previous sections discussed the case where we use one sentence as an input, but we have to input multiple sentences to the downstream model in some tasks.
This section describes our training strategy in such cases.

\begin{figure}[t]
\centering
\includegraphics[scale=0.6]{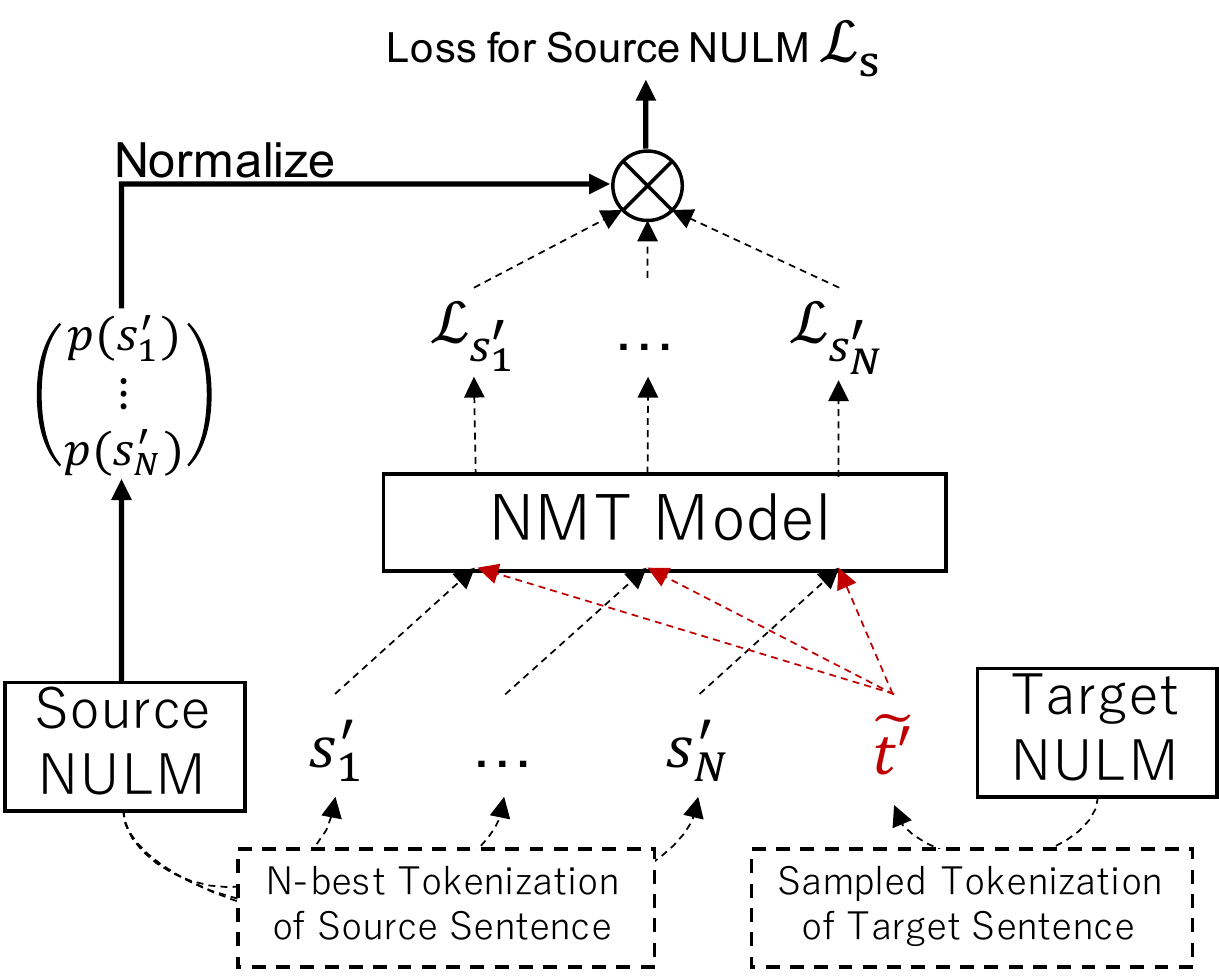}
\caption{
    Overview of the calculation of a tokenization loss $\mathcal{L}_s$ for the source-side NULM in NMT requiring two inputs, source and target sentences $s$ and $t$, respectively.
    The arrows with the continuous line indicate the differentiable path for back-propagation.
}
\label{fgr:model_outline_multi}
\end{figure}

To compute the loss value for training the tokenizer, we consider multiple tokenizations for one sentence and use the sampled tokenization for the others.
For example, in machine translation, we input the source and target sentences to train the downstream model.
The source sentence is the input of the downstream model, and the target sentence is the supervisory signal.
Let $s$ and $t$ be the source sentence and target sentences, respectively, and $s'$ and $t'$ be the corresponding tokenizations.
We update the NULM of the source side using ${\mathcal L}_{s'_n}=q(f(s'_n), \tilde{t'})$, where $\tilde{t'}$ is a sampled tokenization for the target sentence.
We also compute the loss for the NULM of the target side with ${\mathcal L}_{t'_n}=q(f(\tilde{s'}), t'_n)$, where $\tilde{s'}$ is a sampled tokenization.
For training the downstream model, we use sampled tokenizations for all the input sentences.
Thus, we compute ${\mathcal L}_{\tilde{s'}, \tilde{t'}}=q(f(\tilde{s'}), \tilde{t'})$ for the downstream model.
We outline this training process for the NULM of the source side in Figure \ref{fgr:model_outline_multi}, and the training for the target side is explained in the same manner.

\section{Experiment}
\label{sec:experiments}
To validate the applicability of the proposed method to various downstream tasks, we conducted experiments on text classification and machine translation tasks from existing literature.
To compare our method with the existing methods that determine the appropriate tokenization for a specific downstream task, we employ OpTok~\cite{hiraoka2020optimizing} for text classification and DPE~\cite{xuanli2020dynamic} for machine translation.

\begin{table}[t]
\centering
\small
\tabcolsep 3pt
\begin{tabular}{lwc{10mm}wc{10mm}wc{10mm}wc{10mm}}
\hline
            &SP&SP+R&OpTok&Ours \\\hline
Weibo(Zh)*  &92.70&92.79&92.82&\b{93.06} \\
Twitter(Ja)*&85.89&86.51&\b{86.97}&86.92 \\
Twitter(En)*&75.98&77.31&78.52&\b{78.88} \\\hline
Genre(Zh)   &44.19&47.95&48.18 &\b{48.41} \\
Rating(Zh)  &48.96&49.41&49.63&\b{49.76}\\
Genre(Ja)   &46.82&49.84&50.15 &\b{50.79}\\
Rating(Ja)  &52.88&53.43&53.55&\b{53.69}\\
Genre(En)*  &69.93&71.68&\b{71.88}&71.83\\
Rating(En)* &66.42&67.53&67.68&\b{67.90}\\\hline
SNLI*       &75.62&76.75&77.04 &\b{77.05} \\\hline
\end{tabular}
\caption{
    Experimental results on text classification tasks (F1-score).
    SP and R indicate SentencePieec and subword regularization, respectively.
    The highest scores are highlighted in bold.
    Scores of SP+R and OpTok on datasets with * are quoted from \newcite{hiraoka2020optimizing}.
}
\label{tbl:tc_results}
\end{table}

\subsection{Text Classification}
\label{sec:exp_tc}
\paragraph{Settings}
We utilized ten datasets of text classification tasks in three languages.
Weibo(Zh)\footnote{\url{https://github.com/wansho/senti-weibo}}, Twitter(Ja)\footnote{\url{http://www.db.info.gifu-u.ac.jp/data/Data_5d832973308d57446583ed9f}}, and Twitter(En)\footnote{\url{https://www.kaggle.com/c/twitter-sentiment-analysis2}} are sentiment analyses on SNS corpora in Chinese, Japanese, and English, respectively.
Genre and Rating are genre prediction and rating predictions from reviews posted on E-commerce corpora, respectively, in Chinese~\cite{zhang2015daily}\footnote{\url{http://yongfeng.me/dataset/}}, Japanese~\cite{rakuten}, and English~\cite{he2016ups}\footnote{\url{http://jmcauley.ucsd.edu/data/amazon/}}.
In addition, we employed the SNLI corpus~\cite{bowman2015large} to evaluate our method on the setting requiring two sentences as the input.

We focus on SentencePiece (SP)~\cite{kudo2018sentencepiece} and OpTok~\cite{hiraoka2020optimizing} as other tokenizers for comparison with the proposed method.
OpTok is a method to optimize tokenization for text classification by weighting a sentence vector with $N$-best tokenization.
In addition, we trained each model with subword regularization (SP+R)~\cite{kudo2018subword} for fair comparisons.
In subword regularization, we used a sampled tokenization for the training phase and a 1-best tokenization for the inference phase.

For the downstream model, we used a BiLSTM encoder since we followed the experimental configurations of OpTok\footnote{Our implementation is based on the existing code: \url{https://github.com/tatHi/optok}}.
We used SentencePiece to construct vocabulary $V$\footnote{We limit the size of the vocabulary to the half size of the initial vocabulary as well as the training of OpTok.} and initialized unigram probabilities of our NULM.
The initial vocabulary sizes are 16K for Twitter(Ja) and Twitter(En) and 32K for the others.
The number of tokenizations is $N=3$, and the hyperparameters for subword regularization are $\alpha=0.2$ and $k=\infty$.
For the SNLI corpus, the system shares the same NULM for the premise and hypothesis and optimizes the NULM in the manner explained in Section \ref{sec:multi_sentences}.

\paragraph{Results}
Table \ref{tbl:tc_results} presents the experimental results on text classification.
This table indicates that our proposed method surpasses OpTok in eight datasets.
For the other two datasets, the performance of our method is comparable to OpTok.
These results indicate that the proposed method is a better alternative to the existing tokenizers for text classification tasks.
We consider that the difference in the performance is caused by the difference in strategy between the training the downstream models.
OpTok trains the downstream model with a weighted sum of sentence vectors corresponding to $N$-best tokenization with their tokenization probabilities, but it uses $1$-best tokenization in the inference.
This gap might harm the downstream model.
In contrast, since our method trains the downstream model with only one sampled tokenization, the downstream model receives one tokenization in both training and inference consistently.
We consider that this consistency improves the performance.

\subsection{Machine Translation}
\label{sec:exp_mt}
\paragraph{Settings}
For experiments on the machine translation task, we employ IWSLT and WMT corpora on eight language pairs.
We pre-tokenized all the datasets except for the Chinese corpus with Moses Tokenizer\footnote{\url{https://github.com/moses-smt/mosesdecoder}}, and we used jieba\footnote{\url{https://github.com/fxsjy/jieba}} for the Chinese corpus.
We evaluate the performance of each method with detokenized BLEU with SacreBLEU~\cite{post-2018-call} after detokenization.

As a recent tokenizer for machine translation, we compare the proposed method with DPE~\cite{xuanli2020dynamic}, which tokenizes a target sentence, considering the source tokenization, in addition to SentencePiece.
We employed the official implementation of DPE\footnote{\url{https://github.com/xlhex/dpe}} and train the DPE model using SentencePiece tokenization.
In the same as text classification, we used subword regularization as a strong baseline.

For the downstream model, we used Transformer~\cite{vaswani2017attention} implemented in Fairseq~\cite{ott2019fairseq}.
For the IWSLT dataset, we used the small Transformer, and we created the initial vocabulary using SentencePiece with a 16K size of the vocabulary for each language.
For the WMT dataset, we employed Transformer (base), and the size of the vocabulary is 32K.
Similar to the case of text classification tasks, we initialized our NULM with the result of SentencePiece.
The hyperparameters for subword regularization are $\alpha=0.2$ for IWSLT, $\alpha=0.5$ for WMT, and $k=\infty$ for both datasets.
The number of tokenizations for the training of the proposed method is $N=8$ for ISWLT and $N=3$ for WMT.

In the training of NMT with DPE, we applied subword regularization for the source side language, similar to \newcite{xuanli2020dynamic}.
For the proposed method, we prepared three configurations: used our method only for a source side language, only for a target side language, and for both side languages.

\paragraph{Results}
Table \ref{tbl:mt_results} details the performance of each configuration.
This table indicates that the system employing our approach achieves the best performance in most datasets.
The setting where the proposed method is used only for the decoder side succeeds on many datasets.
In contrast, when we use our method for both sides, the performance degrades.
These results imply that it is challenging to optimize the tokenization of source and target languages simultaneously, and it can degrade the performance.
We discuss the simultaneous optimization of source and target languages on NMT in Section \ref{sec:both_enc_dec}.

\begin{table*}[t]
\centering
\small
\tabcolsep 3pt
\begin{tabular}{ccwc{15mm}wc{15mm}wc{15mm}wc{15mm}wc{15mm}wc{15mm}}
\hline
        &Encoder&SP&SP+R&SP+R&Ours&SP+R&Ours \\
        &Decoder&SP&SP+R&DPE&SP+R&Ours&Ours \\\hline
IWSLT14 &De-En&33.79&35.03&35.02&34.90&\b{35.78}&35.13\\
        &En-De&28.09&29.13&29.39&29.56&\b{29.57}&29.30 \\\hline
IWSLT15 &Vi-En&28.70&28.78&28.85&29.34&\b{29.69}&29.44\\
        &En-Vi&30.87&31.60&31.63&31.41&\b{31.74}&31.70\\
        &Zh-En&20.44&21.17&21.38&21.63&21.65&\b{21.89}\\
        &En-Zh&14.40&15.25&15.21&15.45&\b{15.59}&15.31\\\hline
IWSLT17 &Ar-En&29.23&29.39&29.37&29.48&\b{30.04}&29.78 \\
        &En-Ar&15.45&17.75&17.83&\b{18.49}&18.18&18.21 \\
        &Fr-En&37.87&38.43&38.52&\b{38.82}&38.68&38.58 \\
        &En-Fr&37.95&39.83&39.90&40.01&\b{40.08}&39.68 \\\hline
WMT09   &Hu-En&14.84&15.51&\b{15.75}&15.73&15.74&15.60 \\
        &En-Hu&11.02&12.14&12.30&12.30&\b{12.37}&12.33 \\\hline
WMT14   &De-En&31.46&31.89&31.97&\b{32.19}&31.98& 31.90\\
        &En-De&27.10&27.41&27.49&\b{27.62}&27.52&27.44 \\\hline
WMT16   &Ro-En&29.10&31.79&31.80&31.80&\b{31.83}&31.72 \\
        &En-Ro&21.78&24.05&24.29&24.36&\b{24.53}&24.03 \\\hline
\end{tabular}
\caption{
    Results of experiments on machine translation task using IWSLT and WMT corpus (BLEU).
    We show the tokenization method for Encoder and Decoder.
    SP and R mean SentencePiece and subword regularization, respectively.
    The highest scores are highlighted in bold.
}
\label{tbl:mt_results}
\end{table*}

\section{Tokenization as Post-processing}
\paragraph{Settings}
As described in Section \ref{sec:intro}, our proposed method can be applied as post-processing to an already trained model.
In this section, we evaluate the effectiveness of optimizing our tokenizer for the trained model.
Concretely, we trained the NULM with ${\mathcal L}_s$ in Eq.(\ref{eq:total_loss}) without updating the parameters of the downstream model.

We conducted experiments on text classification (Sentiment) and machine translation (IWSLT15) tasks.
We trained the downstream models used in Section \ref{sec:experiments} with subword regularization~\cite{kudo2018subword}.
We trained the models with 30 epochs for text classification and 100 epochs for machine translation.
After the training, we trained only our tokenizer with five epochs using the loss values computed by the trained models.
Moreover, for text classification, we trained OpTok in the same manner as our proposed method as a baseline.

\paragraph{Results}
Table \ref{tbl:post_proc} details the performance of each method.
This table indicates that the proposed method also improves the performance from the base model trained with subword regularization.
The proposed method outperforms OpTok on two datasets of text classification.
Moreover, the proposed method increases the BLEU scores consistently in machine translation.
These results show that the proposed method is useful to improve the performance of the downstream model even if we use a sufficiently trained model as the downstream model.
In other words, since we do not necessarily require training for the proposed method with the downstream model from scratch, our proposed method can be applied to various situations such as the combination with a pre-trained model.

\begin{table}[t]
\centering
\small
\tabcolsep 3pt
\begin{tabular}{cwc{17mm}wc{17mm}wc{17mm}}
\hline
                & SP+R & OpTok & Ours \\\hline
\multicolumn{4}{l}{{\it Sentiment} (F1)} \\
Weibo(Zh)       & 92.69 & \b{93.08} & 92.99 \\      
Twitter(Ja)     & 85.88 & 86.23 & \b{86.28} \\
Twitter(En)     & 77.21 & 77.41 & \b{77.77} \\\hline
\multicolumn{4}{l}{{\it IWSLT15} (BLEU)} \\
Vi-En           & 28.82 & -     & \b{28.91} \\
En-Vi           & 30.48 & -     & \b{30.60} \\
Zh-En           & 21.55 & -     & \b{21.82} \\
En-Zh           & 14.57 & -     & \b{14.83} \\\hline

\end{tabular}
\caption{
    Improvement in the performance by optimizing tokenization as post-processing by OpTok and our method.
    SP+R indicates SentencePiece with subword regularization.
    The highest scores are highlighted in bold.
}
\label{tbl:post_proc}
\end{table}

\section{Discussion}
\subsection{Learning Both Encoder and Decoder}
\label{sec:both_enc_dec}
The results of the machine translation task (Section \ref{sec:exp_mt}) reveal that the performance decreases when we incorporate our method into both the encoder and the decoder sides.
We consider that the cause of this decrease to be the gap in the tokenization strategy between the source and target languages.
In this section, we attempt to make it stable to train our method on both the encoder and decoder sides simultaneously with three possible strategies.\\
\noindent{\bf Enc$\rightarrow$Dec}: We train only the encoder-side NULM in the first 50 epochs, with the decoder-side NULM being frozen; then, we train the decoder-side NULM in the last 50 epochs, with the encoder NULM being frozen.\\
\noindent{\bf Dec$\rightarrow$Enc}: We train our method with the reversed version of the strategy Enc$\rightarrow$Dec strategy.\\
\noindent{\bf Random}: We randomly update either of the NULM on the encoder or that on the decoder sides with at a 0.5 ratio in each mini-batch training.

Table \ref{tbl:exp_both} presents the results of the experiments.
These results indicate that the Enc$\rightarrow$Dec strategy contributes to improving the performance of the simultaneous learning of tokenization on both sides.
In particular, the scores of Vi-En, En-Vi, and Zh-En surpass the best scores reported in Table \ref{tbl:mt_results}, indicating that the Enc$\rightarrow$Dec strategy is effective for the training of our method.
In contrast, the Dec$\rightarrow$Enc strategy decreases the performance in many language pairs.
The performance obtained using the Random strategy is slightly lower than that obtained using the original method (Both).
From these results, we can conclude that it is effective for the machine translation task to learn the tokenization of each side step-by-step, specifically, from the encoder-side to the decoder-side, instead of optimizing both sides simultaneously.

\begin{table}[t]
\centering
\small
\tabcolsep 3pt
\begin{tabular}{cwc{13mm}wc{13mm}wc{13mm}wc{13mm}}
\hline
                & Both  & Enc$\rightarrow$Dec    & Dec$\rightarrow$Enc    & Random \\\hline
Vi-En           & 29.44 & \b{30.22} & 29.47     & 29.37      \\                            
En-Vi           & 31.70 & \b{31.78} & 31.33     & 31.70      \\
Zh-En           & 21.89 & \b{21.99} & 21.82     & 21.66      \\
En-Zh           & 15.31 & \b{15.54} & 14.88     & 15.14    \\\hline

\end{tabular}
\caption{
    Performance of NMT on the IWSLT15 datasets with three strategies for the simultaneous training of our method.
    The scores of Both are taken from the Ours-Ours column in Table \ref{tbl:mt_results}.
    The highest scores are highlighted in bold.
}
\label{tbl:exp_both}
\end{table}

\subsection{Analysis of Tokenization}
\label{sec:tokenization}
\paragraph{Optimized Tokenization}
In this section, we analyze the tokenization obtained using the proposed method on a machine translation task.
Table \ref{tbl:seg_example} presents the comparison of tokenization among SentencePiece, DPE, and our method.
We utilized the IWSLT15 Zh-En corpus for this comparison and tokenized English side sentences using each method.
For our method, we only optimized the English side tokenization.

Table \ref{tbl:seg_example_src} presents a comparison of the tokenization on the source side between SentencePiece and the proposed method.
Our method splits words into smaller segments than SentencePiece, which is the initial tokenization of our method.
For example, our method cuts off the suffix from a stem word, such as splitting ``don'' into ``do-n,'' ``have'' into ``hav-e,'' and ``hours'' into ``hour-s.''

Table \ref{tbl:seg_example_tgt} presentas a comparison of the tokenization on the target side between SentencePiece, DPE, and the proposed method.
Compared to the tokenization on the source side, our method does not split words into tiny units on the target side.
The proposed method exhibits the same tendency of tokenization as DPE, such as splitting the past-suffix ``-ed.''
However, the DPE tokenization contains smaller units than our tokenization; an example of this is the difference in the tokenization for ``away.''

\paragraph{Tokenization Granularity} 
To compare the granularities of each tokenizer, we confirm the number of tokens in the corpus tokenized by each method. 
Table \ref{tbl:analysis_length} presents the ratio of the number of tokens in the training corpus between the initial tokenization (SentencePiece) and the optimized tokenization (DPE and the proposed method). 
In the table, a value greater than 1 indicates an increase in the number of tokens compared to SentencePiece.

The results reveal that the number of tokens in the proposed method increases for the source side tokenization, which means that our method tokenizes a source corpus into small units by splitting morphemes, as shown in Table \ref{tbl:seg_example_src}.

For the tokenization of the target side, the ratio of the number of tokens for the proposed method is slightly smaller than the initial tokenization, other than for the En-Zh pair.
We consider that our method seeks appropriate tokenization to aid in the decoding process while maintaining the granularity of the initial tokenization.
With respect to the translation of the En-Zh pair, our method splits a Chinese sentence into smaller tokens.
Chinese characters contain much more information than English characters, and the number of Chinese tokens in a sentence is smaller than that of English.
We consider that this difference causes the increased tokens on the target side to use the same granularity as the source English corpus.

Compared with the tokenization by the proposed method, the number of tokens for the DPE varies for each language pair.
DPE tokenization is more flexible than our method because DPE employs the Transformer and a special decoding algorithm for tokenization, whereas we simply use a unigaram language model and the Viterbi algorithm.
In addition, DPE tokenizes the target sentence by directly considering the source tokenization by inputting a source sentence to the Transformer.
In contrast, we use the target side NULM trained with the both side information to find the target side tokenization.
Although our tokenization flexibility is limited, our method improves the performance on NMT tasks, as demonstrated in the experimental results.

\begin{CJK}{UTF8}{gbsn}
\begin{table}[t]
\small
\centering
\subfloat[][Tokenziation difference for the source language (En-Zh)]{
\label{tbl:seg_example_src}
\begin{tabular}{wl{4.5mm}|wl{62mm}}
\hline
SP  &  Student s \b{don ' t} \b{have} long \b{hours} of learning . \\
Ours &  Student s \b{do n ' t} \b{hav e} long \b{hour s} of learning . \\\hline
TGT &  学生 在 校 学习 时间 不 长 。 \\\hline
\end{tabular}
}
\\
\subfloat[][Tokenization difference for the target language (Zh-En)]{
\label{tbl:seg_example_tgt}
\begin{tabular}{wl{4.5mm}|wl{62mm}}
\hline
SRC &  引力 与 其它 力 分 隔 开来 \\\hline
SP &  Gra vity \b{separate d} \b{away} from the other force s . \\
DPE &  Gra vity \b{separat ed} \b{a way} from the other force s .\\
Ours &  Gra vity \b{separat ed} \b{away} from the other force s .\\\hline
\end{tabular}
}
\caption{
    Comparison of English tokenization on Zh-En pairs using SentencePiece (SP), DPE, and our method.
    SRC and TGT indicate the tokenization of source and target side, respectively.
    The different tokenization is highlighted in bold.
}
\label{tbl:seg_example}
\end{table}
\end{CJK}

\begin{table}[t]
\centering
\small
\tabcolsep 3pt
\begin{tabular}{cwc{15mm}wc{15mm}wc{15mm}}
\hline
Encoder  &Ours&SP+R&SP+R \\
Decoder  &SP+R&Ours&DPE           \\\hline
\multicolumn{4}{l}{{\it IWSLT14}} \\
De-En&2.5353&0.9992&1.0439 \\
En-De&1.3809&0.9996&0.9923 \\\hline
\multicolumn{4}{l}{{\it IWSLT15}} \\
Vi-En&1.5320&0.9993&1.0428 \\
En-Vi&1.4650&0.9999&0.9923 \\
Zh-En&1.5175&0.9994&0.9907 \\
En-Zh&1.3516&1.4713&1.0346 \\\hline
\multicolumn{4}{l}{{\it IWSLT17}} \\
Ar-En&2.5350&0.9997&0.9952 \\
En-Ar&1.4765&0.9994&0.9945 \\
Fr-En&1.7194&0.9996&1.0001 \\
En-Fr&1.5996&0.9997&0.9935 \\\hline
\end{tabular}
\caption{
    Ratio of the number of tokens between initial tokenization (SentencePiece) and optimized tokenization (DPE and Ours) on the IWSLT corpora.
    SP+R denotes SentencePiece with subword regularization.
}
\label{tbl:analysis_length}
\end{table}

\subsection{Effect of Hyperparameter $N$}
The proposed method updates the NULM using $N$-best tokenized candidates.
In this section, we confirm the effect of the hyperparameter $N$ on the performance for the downstream tasks.

\begin{figure}[t]
\centering
\small
\includegraphics[scale=0.65]{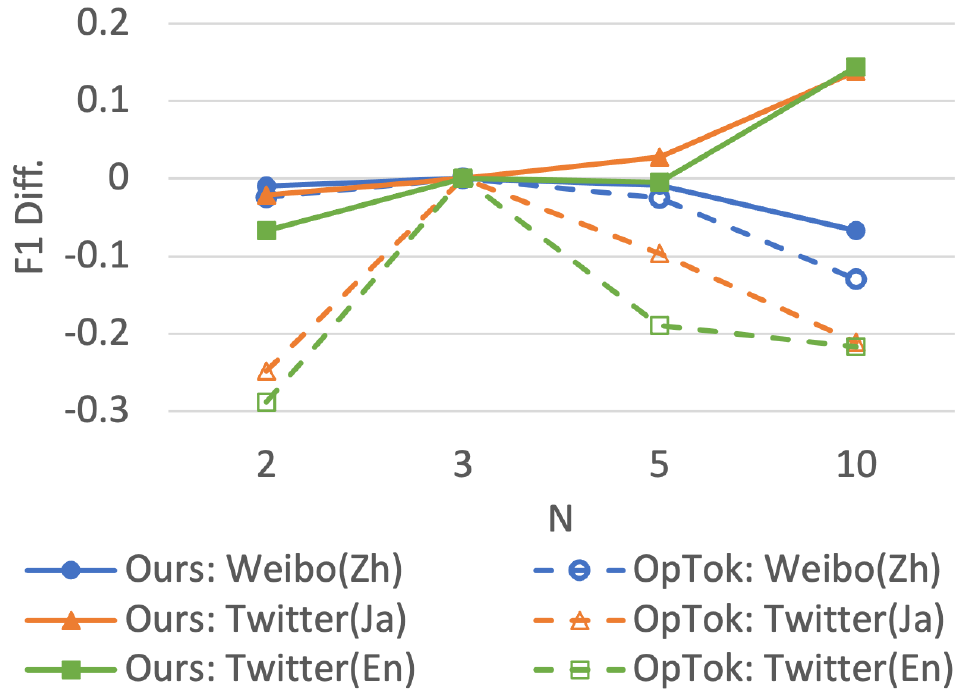}
\caption{
    Difference in performance on text classification tasks against $N$.
}
\label{fgr:various_n_tc}
\end{figure}

\begin{figure}[t]
\centering
\small
\includegraphics[scale=0.65]{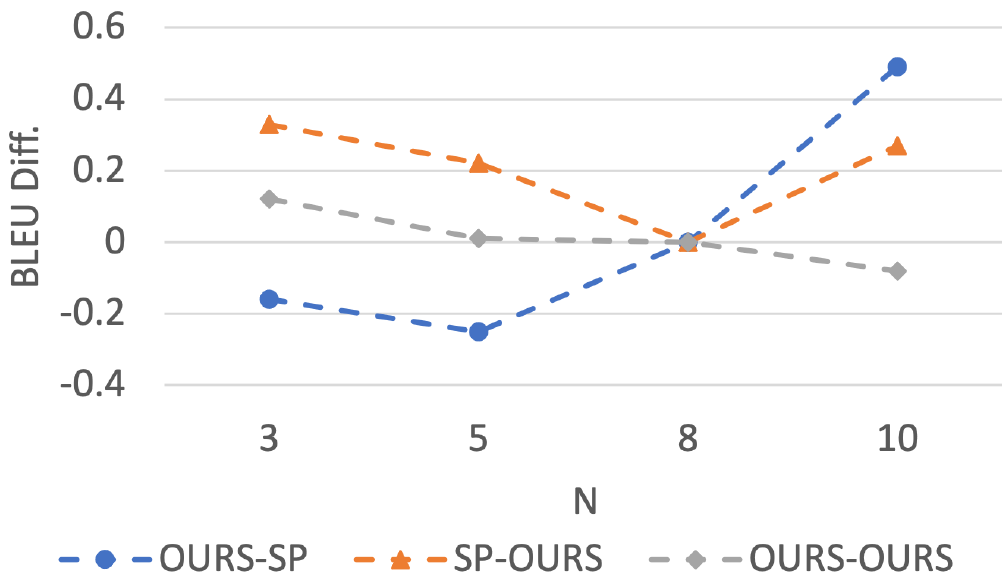}
\caption{
    Difference in performance on machine translation tasks (Vi-En) against $N$.
}
\label{fgr:various_n_mt}
\end{figure}

We conducted experiments on text classification and machine translation with different $N$, as mentioned in Section \ref{sec:experiments}.
Figures \ref{fgr:various_n_tc} and \ref{fgr:various_n_mt} show the results respectively.
In these figures, we illustrate the difference from the performance of the model with the settings used in Section \ref{sec:experiments}, i.e., $N=3$ for text classification and $N=8$ for machine translation.

For the text classification task, we confirm the effect of $N$ on the sentiment analysis datasets.
In Figure \ref{fgr:various_n_tc}, we can observe that the number of $N$ does not have a strong effect on the performance of the proposed method.
In addition, the larger $N$ leads to slightly better performance for the Japanese and English datasets.
In contrast, the performance for the Chinese dataset decreases with a large $N$.
We consider that this occurs because a Chinese sentence has more tokenization candidates than the others, and the optimization of tokenization becomes unstable with larger $N$.

Compared to the existing OpTok method ~\cite{hiraoka2020optimizing} the proposed method is robust to large $N$.
As described in Section \ref{sec:experiments}, our method avoids the gap between training and inference in terms of the weighting strategy.
Because the proposed method uses one sampled tokenization to train the downstream model, the number of $N$ does not affect the text classification performance.
The experiment with various $N$ verifies that our method is superior to \newcite{hiraoka2020optimizing}.

For the machine translation task, we conduct experiments on the Vi-En pair of IWSLT15.
Figure \ref{fgr:various_n_mt} illustrates that the number of $N$ does not have a strong effect on the performance when we use the proposed method solely for the target side (SP-OURS).
When we incorporate our method to the source side (OURS-SP), the performance increases with a large $N$.
We consider that the proposed method is able to seek appropriate tokenization from the large search space when we set a large number as $N$ because the neural encoder of NMT allows various tokenizations for its input.
When we use our method for both the encoder and the decoder (OURS-OURS), the performance decreases slightly with higher $N$.
We consider that optimization of the tokenization of both sides with a large $N$ becomes unstable because tokenization on the source side varies vastly during training.

\section{Related Work}
Many researchers have tackled the problem of optimizing tokenization, especially in the machine translation field.
For statistical machine translation, \newcite{niessen2004statistical} and \newcite{goldwater2005improving} attempted to obtain good tokenization using hand-crafted linguistic information.
Some studies explored appropriate tokenization using alignment information between the source and target languages~\cite{xu2008bayesian, chung2009unsupervised, nguyen2010nonparametric}.

Recent studies have attempted to obtain appropriate tokenization for the downstream task using neural networks.
\newcite{gowda2020finding} analysed the optimal granularity of tokenization on NMT.
\newcite{salesky2020optimizing} proposed Incremental-BPE, which automatically explores the appropriate granularity of BPE tokenization.
They stopped the merge operation of the BPE depending on the loss on a validation split.
\newcite{xuanli2020dynamic} proposesd DPE, which obtains the appropriate tokenization of a target language depending on the tokenization of the source side on the NMT.
Our method is different from DPE in that our method can optimize tokenization considering the parameters of the downstream model.
Moreover, our method can be applied to both the source and target languages of the machine translation task.
\newcite{hiraoka2020optimizing} proposed OpTok, which optimizes the tokenizer and the downstream model on text classification simultaneously.
We extend their idea to be applicable to any downstream task, including machine translation.
Moreover, our method uses a different training strategy to that used in OpTok.
We split the training loss for the downstream loss and tokenization loss, as mentioned in Section \ref{sec:proposed_method}, and the experimental results demonstrate that our strategy is superior to OpTok.

We employ subword regularization to train the downstream model.
\newcite{kudo2018subword} proposed training a model by sampling tokenization with a unigram language model, and \newcite{provilkov2019bpe} modified this idea to use the BPE~\cite{sennrich2016neural} process to yield various tokenizations.
\newcite{hiraoka2019stochastic} applied subword reguralization for text classification tasks.

\section{Conclusion}
We propose a novel method for optimizing tokenization by considering downstream tasks, such as a training corpus and a downstream model.
Our method is the first approach to explore appropriate tokenization for any downstream task.
Experimental results demonstrate that the proposed method achieves higher performance than existing systems with respect to text classification and machine translation tasks.
Because the proposed method is applicable to any architecture using loss for its optimization, we expect our method to improve the performance for other NLP tasks.

\section*{Acknowledgments}
These research results were obtained from the commissioned research by National Institute of Information and Communications Technology (NICT), Japan.

\bibliography{optok_acl2021}
\bibliographystyle{acl_natbib}

\clearpage
\appendix
\section{Detailed Experimental Settings}
For all the experiments, we implement the proposed method using PyTorch, and we run all the experiments on NVIDIA Tesla V100 (16 GiB).
For the initialization of the NULM, we terminate the pretraining when the loss is less than $1\times10^{-7}$ or the training epoch achieves the maximum number (100,000).

\subsection{Text Classification}
In Section \ref{sec:exp_tc}, we use two new datasets in addition to the datasets used in the existing research~\cite{hiraoka2020optimizing}.
We prepare both datasets in the same manner as that used for Genre(En) and Rating(En).

For Genre(Zh) and Rating(Zh), we use 13 genres that contain a sufficient number of reviews and sample 30,000 reviews from each genre, balancing the number of ratings and limiting the number of characters in the review to 100.
The dataset contains 390,000 reviews, and we split it at a ratio of 8:1:1 for training, validation, and testing.
Each review has ratings from 1 to 5 attached by reviewers, and we use the sampled dataset for a rating prediction task and a genre prediction task.

For Genre(Ja) and Rating(Ja), we use 21 genres that contain a sufficient number of reviews and sample 5,000 reviews from each genre and rate, limiting the number of characters in the review to 100.
The dataset contains 525,000 reviews, and we split it at a ratio of 8:1:1 for training, validation, and testing.
Each review has ratings from 1 to 5 attached by reviewers, and we use them for rating and genre prediction tasks.

We conduct experiments on text classification tasks under the same settings as those used in the existing literature~\cite{hiraoka2020optimizing}.
Thus, we conduct a text classification with BiLSTM encoders whose hidden size is 256.
The size of the word embedding is 64, and we set the batch size to 256 and the maximum training epoch to 20.
The pretrained word embeddings are frozen in the training of text classification, and both NULM and the downstream model share word embeddings.

\subsection{Machine Translation}
For experiments on machine translation, we do not freeze the word embeddings.
We empirically find that the training becomes unstable when word embeddings are shared between NULM and the downstream model without freezing.
Therefore, we prepare different word embeddings for NULM and the downstream model, called the Transformer.
We set the word embedding size to 64 for the NULM. 
We make the mini-batch by specifying the number of maximum tokens, and we set it to 1,000.
The maximum number of training epochs is 100 for all the experiments, and we average the parameters of the last 10 epochs for evaluation.

\begin{table*}[t]
\centering
\small
\begin{tabular}{ll}
\hline
Method & Tokenization \\\hline
SentencePiece    &   This didn ' t have full \b{seasons} . I expected \b{more} than 6 \b{episodes} on this because I \b{liked} this series .\\\hline
\multicolumn{2}{l}{{\it Genre (Gold: Movies\_and\_TV)}} \\
OpTok  &   This didn ' t have full \b{seasons} . I expected \b{more} than 6 \b{episode s} on this because I \b{liked} this series .\\
Ours  &   This didn ' t have full \b{season s} . I expected \b{more} than 6 \b{episode s} on this because I \b{liked} this series .\\\hline
\multicolumn{2}{l}{{\it Rating (Gold: 1)}} \\
OpTok  &   This didn ' t have full \b{seasons} . I expected \b{mor e} than 6 \b{episodes} on this because I \b{like d} this series .\\
Ours  &   This didn ' t have full \b{seasons} . I expected \b{more} than 6 \b{episodes} on this because I \b{liked} this series .\\\hline
\end{tabular}
\caption{
    Tokenization examples on the text classification task.
    Bold highlights the difference of tokenization among methods.
}
\label{tbl:seg_tc}
\end{table*}

\section{Tokenization for Text Classification}
We present examples for the tokenization on text classification tasks, Genre/Rating(En), in Table \ref{tbl:seg_tc}.
As both Genre and Rating datasets are created from the same review corpus, we can confirm whether each method can tokenize a sentence depending on the task, which might be a genre prediction or a rating prediction.
We compare the tokenization by SentencePiece, OpTok, and our method.

The tendency for tokenization by our method is similar to that by OpTok because our method is based on OpTok.
In the example, both OpTok and our method split a suffix ``s'' from ``episodes'' only on the genre prediction task.
This example implies that both methods yield tokenization that includes task-specific word such as ``episode'' for the ``Movies\_and\_TV'' genre.
In addition, our method tokenizes ``seasons'' into ``season-s,'' which is also related to the movie genre.

With respect to the rating prediction, our method does not split ``liked'' into ``like,'' which might be helpful for predicting ratings, whereas OpTok does.
We consider that our method uses the original word to distinguish the verb from the adjective/adverb ``like.''

\end{document}